\documentclass[letterpaper]{article} 
\usepackage{aaai24}  
\usepackage{times}  
\usepackage{helvet}  
\usepackage{courier}  
\usepackage[hyphens]{url}  
\usepackage{graphicx} 
\usepackage{natbib}  
\usepackage{caption} 
\usepackage{algorithm}
\usepackage{algorithmic}
\usepackage{makecell}
\usepackage{multirow}
\usepackage{booktabs}
\usepackage{tabularx}
\usepackage{array}
\newcommand{\ie}{\textit{i.e.}\ }
\newcommand{\eg}{\textit{e.g.}\ }
\newcommand{\wrt}{\textit{w.r.t.}\ }
\usepackage{newfloat}
\usepackage{listings}
\DeclareCaptionStyle{ruled}{labelfont=normalfont,labelsep=colon,strut=off} 
\floatstyle{ruled}
\newfloat{listing}{tb}{lst}{}
\floatname{listing}{Listing}
\title{Empowering CAM-Based Methods with Capability to Generate \\Fine-Grained and High-Faithfulness Explanations}
\author{
    Changqing Qiu\textsuperscript{\rm 1},
    Fusheng Jin\textsuperscript{\rm 1}\thanks{Corresponding author.},
    Yining Zhang\textsuperscript{\rm 2}
}
\affiliations{
    \textsuperscript{\rm 1}Beijing Institute of Technology\\
    \textsuperscript{\rm 2}Peking University\\
    changqing\_qiu@bit.edu.cn, jfs21cn@bit.edu.cn,
    zhangyining@stu.pku.edu.cn

}

\usepackage{bibentry}

\begin{document}

\maketitle

\begin{abstract}
Recently, the explanation of neural network models has garnered considerable research attention. 
In computer vision, CAM (Class Activation Map)-based methods and LRP (Layer-wise Relevance Propagation) method are two common explanation methods. 
However, since most CAM-based methods can only generate global weights, they can only generate coarse-grained explanations at a deep layer. 
LRP and its variants, on the other hand, can generate fine-grained explanations. 
But the faithfulness of the explanations is too low. 
To address these challenges, in this paper, we propose FG-CAM (Fine-Grained CAM), which extends CAM-based methods to enable generating fine-grained and high-faithfulness explanations. 
FG-CAM uses the relationship between two adjacent layers of feature maps with resolution differences to gradually increase the explanation resolution, while finding the contributing pixels and filtering out the pixels that do not contribute. 
Our method not only solves the shortcoming of CAM-based methods without changing their characteristics, but also generates fine-grained explanations that have higher faithfulness than LRP and its variants.
We also present FG-CAM with denoising, which is a variant of FG-CAM and is able to generate less noisy explanations with almost no change in explanation faithfulness. 
Experimental results show that the performance of FG-CAM is almost unaffected by the explanation resolution. 
FG-CAM outperforms existing CAM-based methods significantly in both shallow and intermediate layers, and outperforms LRP and its variants significantly in the input layer.
Our code is available at \url{https://github.com/dongmo-qcq/FG-CAM}.
\end{abstract}

\section{Introduction}
In recent years, deep learning has been widely used in many areas. As a result, it is increasingly important to explain why the network makes a specific decision. 
There has been extensive research on the interpretability of deep learning, especially in some important areas such as medicine~\shortcite{A25,A20,A22,A24,A23} and autonomous driving~\shortcite{A21,A26,A27}.
In computer vision, a very large number of methods have been proposed, such as CAM-based methods~\shortcite{A3,A14,A36,A6,A35,A15,A34,A5,A37,A7,A2,A4,A28,A1,A39}, LRP~\shortcite{A8} and its variants~\shortcite{A9,A10}, and other methods~\shortcite{A30,A16,A12,A33,A32,A38}.

\begin{figure}[t]
	\centering
	\includegraphics[width=1.0\linewidth]{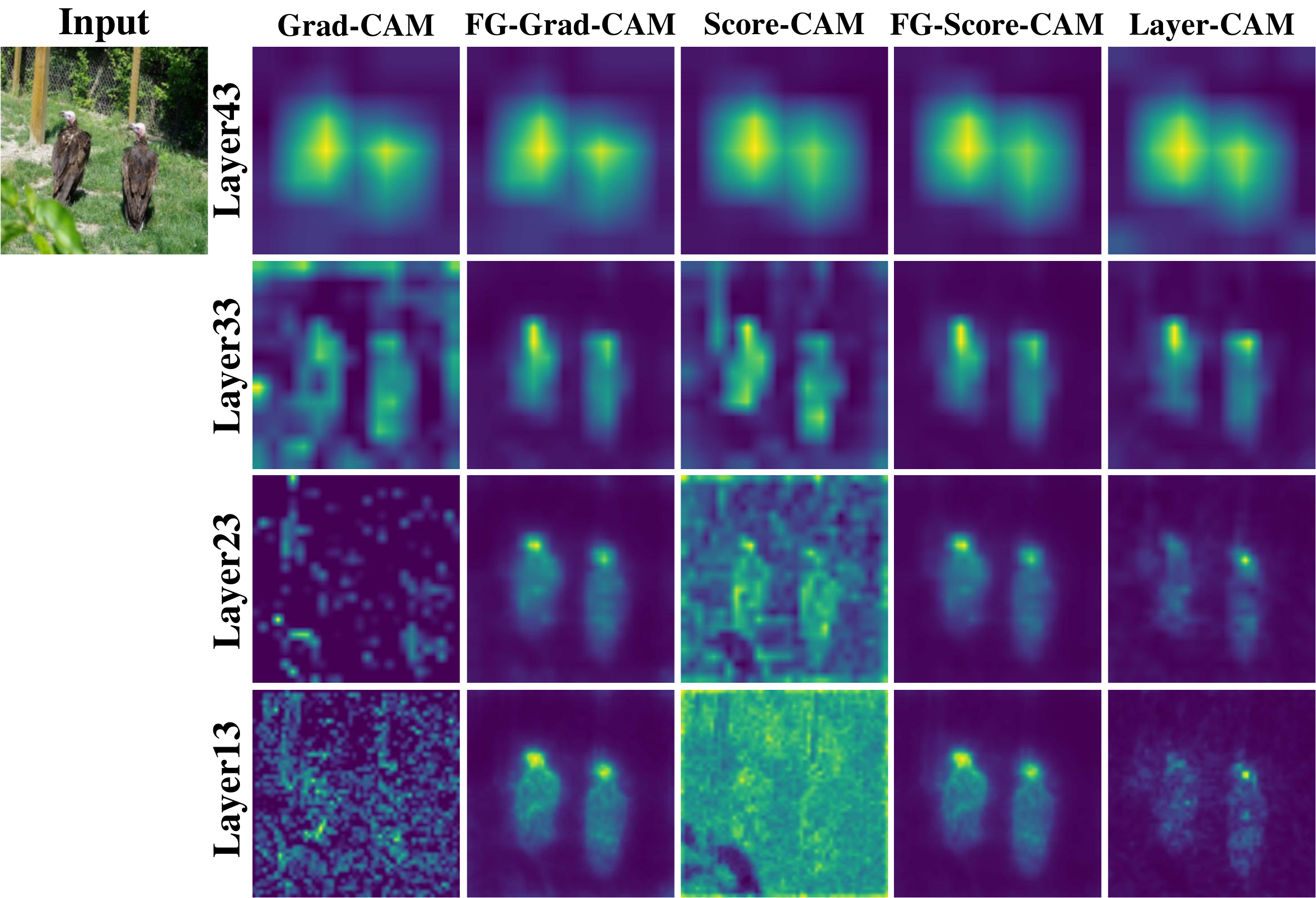}
	\caption{Visualization results of FG-CAM and other CAM-based methods on VGG-16 with batch normalization. Layer43 represents model.features[43]. Pixels with higher brightness are more important.}
	\label{fig:1}
\end{figure}

CAM method~\shortcite{A1} generates an explanation through a linearly weighted combination of the feature maps of the last convolutional layer.
Although CAM method can locate important regions in input image, it has great restrictions on the model structure.
Therefore, as an extension of CAM, Grad-CAM~\shortcite{A2}, Score-CAM~\shortcite{A4} and other CAM-based methods have been proposed. 
However, before Layer-CAM~\shortcite{A6}, all CAM-based methods could only be applied to the final convolutional layer of the model. If the target layer became shallow, the faithfulness (measuring how accurately the explanation reveals the model decision) of the explanations would be significantly reduced. Thus, only coarse-grained explanations could be generated, providing only a rough location of the target object. This limitation restricted performance on tasks requiring pixel-accurate object locations~\cite{A6}.
Layer-CAM, on the other hand, although outperforming other CAM-based methods on shallow layers, suffers from shattered gradient problem~\cite{A13} as the model gets deep and the target layer becomes shallow, which can lead to problems such as making explanation noisy and reduced faithfulness (not as dramatically as other CAM-based methods, but still not negligible).
As shown in Figure~\ref{fig:1}, as the target layer becomes shallower, Grad-CAM and Score-CAM diminish the explanation accuracy significantly and even become unusable (unable to extract contributing pixels or filter out non-contributing pixels). 
The explanation of Layer-CAM has become more and more noisy and less accurate, while the FG-Grad-CAM and FG-Score-CAM still show good performance. 	
LRP and its variants, which are based on the Deep Taylor Decomposition~\cite{A11}, backpropagate the relevance score from output layer to input layer through specific rules to obtain the contribution of each pixel in the input image. 
They can generate explanations at the same resolution as the input image, but their faithfulness is low.

In order to empower CAM-based methods with capability to generate fine-grained and high-faithfulness explanations without changing their characteristics, this paper proposes a novel method: FG-CAM. 
Different from other methods which are directly applied to shallow layers to obtain high-resolution explanations, FG-CAM uses the relationship between two adjacent layers of feature maps with resolution differences to gradually increase the resolution of explanations produced on the last convolutional layer, while finding the contributing pixels and filtering out pixels that do not contribute.
FG-CAM not only solves the problem that CAM-based methods can only provide rough explanations without changing their characteristics, but also generates fine-grained explanations that have higher faithfulness than LRP and its variants. 
This paper also presents FG-CAM with denoising, which is able to generate less noisy explanations with almost no change in explanation faithfulness.

To the best of our knowledge, our method achieves the SOTA in shallow and intermediate layers as well as in the input layer (different layers represent different explanation resolutions). 
The contributions of this paper are as follows.

\begin{enumerate}
	\item This paper proposes FG-CAM, a novel explanation method, which can generate fine-grained explanations with high faithfulness in shallow and intermediate layers as well as in the input layer. FG-CAM solves the shortcoming of CAM-based methods without changing their characteristics, and achieves generality. Different from other CAM-based methods, the performance of FG-CAM is almost unaffected by the explanation resolution.
	\item We also present FG-CAM with denoising, which is a variant of FG-CAM and can generate less noisy explanations with almost no change in explanation faithfulness.
	\item We have verified qualitatively and quantitatively that FG-CAM performs significantly better than Layer-CAM (currently the best-performing CAM-based method for shallow layers), Grad-CAM, and Score-CAM in both shallow and intermediate layers. It is proved that whether FG-CAM is applied or not makes a very large difference.
	\item Through experiments, we verify that FG-CAM can generate explanations with high faithfulness in the input layer, which is not possible for other CAM-based methods, and also outperforms LRP and its variants significantly.
	
\end{enumerate}

\section{Related Work}

\subsection{CAM-Based Methods}

CAM-based methods generate an explanation through a linearly weighted combination of the feature maps of a certain layer (usually the last convolutional layer). Consider a layer $l$ in a model $f$, $A_l^i$ represents the feature map of the $i$th channel output by the layer, then the explanation of the CAM-based methods for class $c$ can be defined as:

\begin{equation}
	L_{CAM}^c=ReLU(\sum_{i}w_c^iA_l^i)
	\label{eq:important}
\end{equation}

In original CAM, $w_c^i$ is the connection weight between classification layer and global average pooling layer which is applied in the final convolution layer. 
It has significant limitations on the model structure.

Grad-CAM uses the average gradient of the output \wrt the feature map as the weight, solving the model structure dependency problem of CAM. 
Grad-CAM is equivalent to CAM when the target layer is followed by one and only one fully connected layer. Other gradient-based CAM methods, such as Grad-CAM++~\shortcite{A3} and XGrad-CAM~\shortcite{A14}, will not be discussed in this paper.

Different from gradient-based CAM methods, Score-CAM uses upsampled and normalized feature maps as masks of the image. 
The image perturbed by the masks is input into the model, and the weights of the feature maps are obtained according to the output. 
Score-CAM gets rid of the dependence on the gradient and shows very good performance. 
However, Score-CAM still can not be used at shallow layers.
We use Grad-CAM and Score-CAM to verify that FG-CAM can be used on the CAM-based methods based on both score weight and gradient weight.

Unlike other CAM-based methods where one feature map corresponds to one global weight, Layer-CAM uses pixel-level weights and uses the positive gradient of the output \wrt the pixel as the weights (\ie $w_c^i=ReLU(\frac{\partial y^c}{\partial A_l^i}$) to address the problem that CAM-based methods can not be used at shallow layers.
Layer-CAM uses fine-grained weights and can generate high-resolution explanations at shallow layers, which is not possible for the majority of CAM-based methods. However, as the network gets deeper, the gradients become noisy and discontinuous~\cite{A15}.

\subsection{LRP and Its Variants}
LRP is another explanation method, which backpropagates the relevance score from the output layer to the input layer through specific rules to obtain the contribution of input image pixels to the model output.

The general propagation rule for LRP is the $z^+$ rule and $z^\beta $ rule. 
For the output layer, LRP uses the model's output as relevance scores and sets the scores for non-target classes to 0.
After propagating relevance score to the input layer, LRP determines the contribution of each pixel in the input image based on its relevance score.
Therefore, LRP can generate explanations with the same resolution as input image.
While LRP can generate fine-grained explanations, the results are not class-discriminative~\cite{A9}. 
To address this problem, CLRP~\shortcite{A9} and SGLRP~\shortcite{A10} were proposed, which differ from LRP only in their definition of the relevance score for the output layer.
However, CLRP and SGLRP, which over-remove pixels that contribute to other classes, often result in important parts of the target class being removed (even removing the entire object).
It can significantly affect the correctness of the explanation and may lead people in the wrong direction. 
Therefore, the explanations generated by LRP, CLRP, and SGLRP have low faithfulness.

\begin{figure*}
	\centering
	\includegraphics[width=0.8\linewidth]{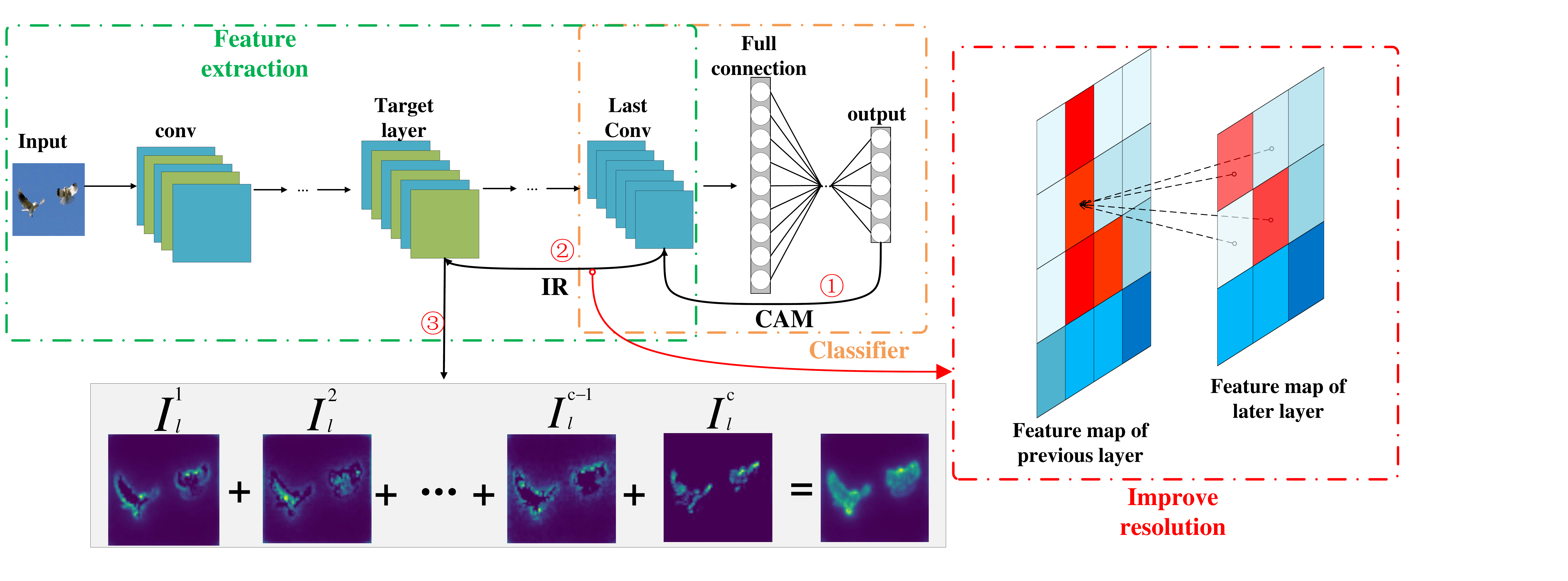}
	\caption[width=0.8\linewidth]{FG-CAM pipeline. In step 1, the explanation components are calculated using a CAM-based method in last convolutional layer. In step 2, improve resolution of the explanation components. Finally, in step 3, generate a fine-grained explanation.}
	\label{fig:3}
\end{figure*}

\section{Our Method}

\subsection{FG-CAM}
CAM-based methods usually assign a global weight to a feature map (\ie every pixel within the feature map has the same weight). 
However, in shallow layers, the contribution of each pixel in a feature map can be very different (\eg contains pixels of both the target class and other classes simultaneously).
The global weight cannot filter out these irrelevant pixels and highlight important pixels.
Thus the generated explanation cannot focus attention on the important regions (\eg Figure \ref{fig:1}, explanations of Score-CAM and Grad-CAM on Layer13).
Therefore, CAM-based methods can only be applied in the last convolutional layer to generate explanations that have very low resolution even though highlighting important regions (\eg for VGG-16 with batch normalization, the explanation only has $7\times7$ resolution, which is much lower than $224\times224$ of the input image).
And we need to be able to determine the importance of each pixel in order to generate a fine-grained explanation.

Layer-CAM proposes fine-grained weights so that it can be used directly at shallow layers to generate high-resolution explanations.
However, the explanations generated by Layer-CAM are very noisy, and almost all CAM-based methods fail to produce fine-grained weights.
Therefore, in order to achieve generality, we address the problem that CAM-based methods can only generate coarse-grained explanations from another perspective, which is to improve the explanation resolution, while finding the contributing pixels and filtering out the pixels that do not contribute (\ie make the explanation fine-grained), rather than trying to use CAM-based methods directly on the shallow layers.  

For the final explanation, it is difficult for us to make it fine-grained because we cannot easily obtain the relationship between the correct fine-grained explanation and the current explanation. 
In other words, we cannot accurately determine the importance of each pixel in the fine-grained explanation through the current explanation. 
However, for feature maps, this is very easy to implement.
For two adjacent layers, the feature maps of previous layer usually have more details and higher resolution, and the feature maps of the later layer are calculated from the feature maps of the previous layer (\ie we can easily obtain the relationship between the feature maps of adjacent layers).
Therefore, if we have the importance of each pixel in the feature map of later layer, then we can determine the importance of the pixels in the previous layer's feature map through the relationship between the two adjacent layer's feature maps (\eg based on computational relationships, we can allocate the importance of pixels from the later layer to pixels in the previous layer using such as gradients, specific rules, and so on).

Through the above method, we obtain the feature maps with higher resolution and more detail, and filter out non-contributing pixels within them.
By performing this layer by layer, we can obtain the importance of each pixel in the feature map with higher resolution of the target layer.

Therefore, we can first use CAM-based methods to obtain the importance of each pixel in the last convolutional layer (without combining them to get the final explanation), which can be viewed as the explanation components.
Then we use the relationship between the two adjacent layers' feature maps to improve the resolution of the explanation components.
Finally, we sum the processed components to obtain fine-grained explanations. 
The pipeline of FG-CAM is shown in Figure~\ref{fig:3}. 

Consider a model $f$, the last convolutional layer of $f$ is $L$ (the shape of output is $C\times W \times H$), the target layer is $l$ (the shape of output is $c\times w\times h$).
The FG-CAM explanation, written $L_{FG-CAM}$, can be calculated as follows.

\textbf{Step 1: calculate the explanation components in $L$.}
\begin{equation}
	w^{(C\times 1\times 1)}=CAM(A_L^{(C\times W\times H)})
	\label{eq:important}
\end{equation}
\begin{equation}
	I_L^{(C\times W\times H)}=w^{(C\times 1\times 1)}A_L^{(C\times W\times H)}
	\label{eq:important}
\end{equation}
where $A_L$ is the output of layer $L$, $w$ denotes global weights of feature maps in layer $L$ and can be obtained by any CAM-based methods.
$I_L$ denotes the explanation components in layer $L$. 

\textbf{Step 2: improve the resolution of $I_L$ to obtain $I_l$.}
\begin{equation}
	I_l^{(c\times w\times h)}=IR(I_L^{(C\times W\times H)})
	\label{eq:important}
\end{equation}
where $IR()$ denotes improving the resolution of the explanation components layer by layer while finding the contributing pixels and filtering out the non-contributing pixels. In this paper, we implement IR based on the $z^+$ and $z^\beta$ rule.

Consider a model, $i$ represents the $i$th pixel in layer $l$, $j$ represents the $j$th pixel in layer $l+1$, and $w_{ij}^ {(l)}$ represents the weight of connection between $i$ and $j$, and the importance of $i$ and $j$ is $I_i^l$ and $I_j^{l+1}$ respectively. Then, in this paper, the $z^+$ rule can be defined as:
\begin{equation}
	I_{ij}^{l}=\frac{w_{ij}^{+(l)}x_i^l}{\sum_{i'}{w_{i'j}^{+(l)}x_{i'}^l}}I_j^{l+1}
	\label{eq:important}
\end{equation}
\begin{equation}
	I_i^l=\sum_{j}I_{ij}^l
	\label{eq:important}
\end{equation}
where $w_{ij}^{+(l)}=max(w_{ij}^{(l)},0)$, $x_i^l$ is the pixel value of $i$ in layer $l$.
The $z^\beta $ rule can be defined as:

\begin{equation}
	I_{ij}^l=\frac{w_{ij}^{(l)} x_i^l-w_{ij}^{+(l)} b_i^l-w_{ij}^{-(l)} h_i^l}{\sum_{i'}w_{i'j}^{(l)} x_{i'}^l-w_{i'j}^{+(l)} b_{i'}^l-w_{i'j}^{-(l)} h_{i'}^l} I_j^{l+1}
	\label{eq:important}
\end{equation}
where $w_{ij}^{-(l)}=min(w_{ij}^{(l)},0)$, interval $[b_i^l, h_i^l]$ is the domain of the pixel value $x_i^l$. 

\textbf{Step 3: sum the $I_l$ to obtain $L_{FG-CAM}$.}
\begin{equation}
	L_{FG-CAM}=ReLU(\sum_{i}^{c}I_l^i)
	\label{eq:important}
\end{equation} 

Therefore, different from trying to apply CAM-based method directly to shallow layers, FG-CAM solves the problem that CAM-based methods cannot generate fine-grained explanations while maintaining the good performance and characteristics of CAM-based methods, achieving good generality. 
Even if new CAM-based methods are proposed in the future, FG-CAM can still be used to generate fine-grained explanations based on these methods.
In addition, it can be found that FG-CAM can generate explanations with the same resolution as the input image, which is impossible with CAM-based methods.

\subsection{FG-CAM with Denoising}
Due to the global weight, the explanations generated by the existing CAM-based methods are likely to be noisy. 
These noises will spread as the explanation resolution increases, resulting in many background pixels that have positive contributions (although their importance is very small compared to the target class pixels, they may also cause interference).
To solve the problem, we propose FG-CAM with denoising, a variant of FG-CAM with one more step. 
After getting $I_L$, we first reshape $I_L$ to $I_L^{(C\times K)}$, and for each row in matrix $I$, subtract the mean of that row.
Secondly, we perform singular value decomposition on $I_L^{(C\times K)}$, and then select the top 10$\%$ singular values to reconstruct $I_L^{(C\times K)}$, finally reshape $I_L^{(C\times K)}$ to $I_L$. 
Then, we use the new $I_L$ as the input to $IR()$. 
FG-CAM with denoising can generate explanations that focus more attention on the target class and have less noise.

\section{Experiment}
In this section, we compare FG-CAM with other methods through qualitative and quantitative experiments. FG-Score-CAM denotes FG-CAM that uses Score-CAM to obtain the global weights, and so for FG-Grad-CAM.
We compare FG-Grad-CAM, FG-Score-CAM with Grad-CAM, Score-CAM, and Layer-CAM to verify that FG-CAM does not significantly reduce the faithfulness of explanation as the target layer becomes shallower, as is the case with CAM-based methods, therefore solving the problem that CAM-based methods cannot generate fine-grained explanations, achieving SOTA.
We also verify that FG-CAM has better performance than LRP, CLRP, and SGLRP in the input layer. 

In the following experiments, we use pre-trained VGG-16~\shortcite{A19} with batch normalization from PyTorch model zoo as a base model. ILSVRC2012 val~\shortcite{A29} is used as the data set. For each image, we resize it to $(224\times 224 \times 3)$, convert it to range [0, 1], then normalize it using mean vector [0.485, 0.456, 0.406] and standard deviation vector [0.229, 0.224, 0.225], and no further pre-processing beyond that. Layer13, Layer23, and Layer33 represent the 2nd, 3rd, and 4th max pooling layer, respectively.
FG-CAM\textsuperscript{D} represents FG-CAM with denoising.
For FG-CAM, LRP and its variants, we use the $z^\beta $ rule in input layer and the $z^+$ rule in other layers.
The blurred image is realized by Gaussian blur, that is, $\widetilde{I}=gaussian\_{blur2d}(I, ksize, sigma)$, where $I$ is the original image, $\widetilde{I}$ is the blurred image. In this paper, we set $ksize=51$, $sigma=50$. 
The experimental results of FG-CAM have been bolded.

\subsection{Comparison with CAM-Based Methods}
In this section, we verify that FG-CAM can generate fine-grained explanations with high faithfulness and has better performance in both shallow and intermediate layers.

\begin{figure}[t]
	\centering
	\includegraphics[width=1.0\linewidth]{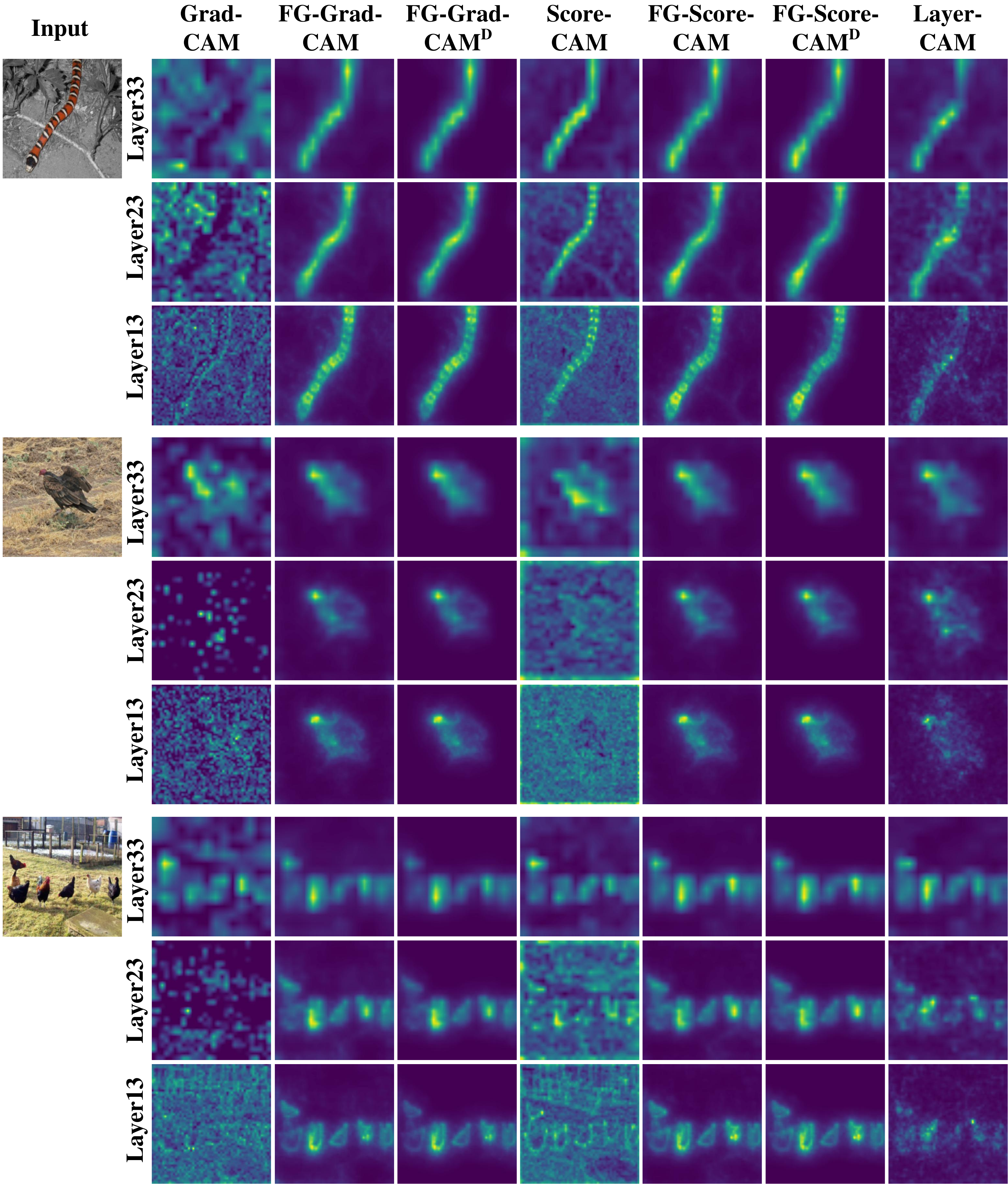}
	
	\caption{Qualitative comparisons of FG-CAM with other methods results. The results of FG-CAM are less noisy.}
	\label{fig:4}
\end{figure}

\normalsize {\textbf{Class Discriminative Visualization}}. 
We compare the visual explanation of each method in shallow and intermediate layers. 
As shown in Figure~\ref{fig:4}, as the target layer becomes shallower, other methods diminish the explanation accuracy significantly, or even become unusable (can not filter out pixels that do not contribute), while FG-CAM still performs very well.
At the same resolution, FG-CAM performs significantly better than other methods.
The higher the explanation resolution, the bigger the gap between other methods and FG-CAM.
Therefore, when generating high-resolution explanations, FG-CAM can still accurately extract the pixels with contribution and filter out non-contributing pixels. 
The explanations generated by FG-CAM with denoising have the least noise.

\normalsize {\textbf{Evaluation on Faithfulness}}. 
Faithfulness is an important characteristic to evaluate explanation methods, which measures whether the explanations correctly highlight regions related to the model's decision.
For the Average Drop/Increase metric, we retain only the most important 50$\%$ pixels of the input image, and then evaluate the faithfulness by Average Drop (A.D.) and Average Increase (A.I.) as \citeauthor{A15,A7,A4}
The A.D. is expressed as $\sum_{i}^{N}\frac{\max(0,y_i^c-o_i^c)}{y_i^c}\times \frac{100}{N}$, and the A.I. is expressed as $\sum_{i}^{N}\frac{sign(y_i^c<o_i^c)}{N}\times 100$
, where $N$ is the number of images, $y_i^c$ is the output softmax value for class $c$ on image $i$ , and $o_i^c$ is the output softmax value for class $c$ on the image after removing 50$\%$ of the pixels, $sign()$ indicates a function that returns 1 if the input is true and 0 otherwise.
If the explanations correctly highlight important regions, Average Drop will have a low value and Average Increase will have a high value.
We randomly selected 2000 images from the ILSVRC2012 val for evaluation as \citeauthor{A4,A15}
To remove 50$\%$ of the pixels, we replace the least important 50$\%$ with blurred pixels. The experimental result is reported in Table~\ref{tab:1}.
\begin{table}
	\centering
	\normalsize
		\begin{tabularx}{\linewidth}{@{\hspace{0em}}l@{\hspace{0.4em}} XXXXXX}
			\toprule
			& \multicolumn{2}{c}{Layer33} & \multicolumn{2}{c}{Layer23} & \multicolumn{2}{c}{Layer13} \\
			\midrule
			Method & A.D. & A.I. & A.D. & A.I. & A.D. & A.I. \\
			\hline
			Grad-CAM & 44.53 & 18.85 & 72.89 & 5.10 & 83.80 & 3.25 \\
			\hline
			FG-Grad-CAM & \textbf{18.69} & \textbf{34.90} & \textbf{19.55} & \textbf{33.45} & \textbf{19.65} & \textbf{32.35} \\
			\hline
			FG-Grad-CAM\textsuperscript{D} & \textbf{18.00} & \textbf{33.85} &
			 \textbf{17.75} & \textbf{33.55} & \textbf{17.90} & \textbf{34.00} \\
			\hline
			Score-CAM & 32.63 & 21.60 & 51.24 & 13.80 & 57.06 & 12.00 \\
			\hline
			FG-Score-CAM & \textbf{21.13} & \textbf{29.15} & \textbf{22.63} & \textbf{27.15} & \textbf{23.50} & \textbf{25.85} \\
			\hline
			FG-Score-CAM\textsuperscript{D} & \textbf{19.19} & \textbf{31.70} & \textbf{18.99} & \textbf{32.00} & \textbf{19.24} & \textbf{30.50} \\
			\hline
			Layer-CAM & 24.25 & 26.80 & 27.32 & 24.50 & 30.76 & 24.40 \\
			\bottomrule
	\end{tabularx}
	\caption{The results of the faithfulness evaluation (lower is better for A.D., higher is better for A.I.).}
	\label{tab:1}
\end{table}

\begin{table*}[t]
	\centering
		\begin{tabular}{lcccccccc}
			\toprule
			&&Grad-CAM&\makecell{FG-Grad-\\CAM}&\makecell{FG-Grad-\\CAM\textsuperscript{D}}&Score-CAM&\makecell{FG-Score-\\CAM}&
			\makecell{FG-Score-\\CAM\textsuperscript{D}}&Layer-CAM\\
			\midrule
			\multirow{3}{*}{Layer33} & Insertion &0.43920&0.60214&0.60247&0.51428&0.58202&0.59716&0.56566 \\
			& Deletion&0.15392&0.09640&0.11108&0.11988&0.10196&0.11198&0.09997 \\
			&Over-all&0.28528&\textbf{0.50574}&\textbf{0.49139}&0.39440&\textbf{0.48006}&\textbf{0.48518}&0.46569 \\
			\hline
			\multirow{3}{*}{Layer23} &Insertion&0.30033&0.59217&0.60248&0.39734&0.57210&0.59676&0.54096\\
			&Deletion&0.19113&0.09459&0.11378&0.14254&0.09985&0.11403&0.09333\\
			&Over-all&0.10920&\textbf{0.49758}&\textbf{0.48870}&0.25480&\textbf{0.47225}&\textbf{0.48273}&0.44763\\
			\hline
			\multirow{3}{*}{Layer13}
			&Insertion&0.23266&0.58630&0.59976&0.36485&0.56570&0.59341&0.51348\\
			&Deletion&0.16850&0.09256&0.11421&0.15037&0.09774&0.11435&0.08469\\
			&Over-all&0.06416&\textbf{0.49374}&\textbf{0.48555}&0.21448&\textbf{0.46796}&\textbf{0.47906}&0.42879\\
			\bottomrule
	\end{tabular}
	\caption{Results of deletion/insertion tests. The Over-all score (higher is better) shows that FG-CAM outperforms other methods.}
	\label{tab:2}
\end{table*}

\begin{table*}
	\centering
	\small
		\begin{tabular}{lccccccc}
			\toprule
			&Grad-CAM&FG-Grad-CAM&FG-Grad-CAM\textsuperscript{D}&Score-CAM&FG-Score-CAM&FG-Score-CAM\textsuperscript{D}&Layer-CAM\\
			\midrule
			Layer33&0.32417&\textbf{0.51458}&\textbf{0.66736}&0.35646&\textbf{0.57161}&\textbf{0.68256}&0.49272\\
			\hline
			Layer23&0.31645&\textbf{0.50995}&\textbf{0.65744}&0.28588&\textbf{0.56192}&\textbf{0.67233}&0.49854\\
			\hline
			Layer13&0.30760&\textbf{0.50861}&\textbf{0.65472}&0.29219&\textbf{0.56001}&\textbf{0.66959}&0.48238\\
			\bottomrule
	\end{tabular}
	\caption{Comparative evaluation on localization ability (higher is better).}
	\label{tab:3}
\end{table*}

\begin{table*}[t]
	\centering
	\begin{tabular}{lccccccc}
		\toprule
		&LRP&CLRP&SGLRP&FG-Grad-CAM&\makecell{FG-Grad-CAM\textsuperscript{D}}&FG-Score-CAM&\makecell{FG-Score-CAM\textsuperscript{D}}\\
		\midrule
		A.D.&25.48&30.64&40.62&\textbf{16.81}&\textbf{16.13}&\textbf{20.28}&\textbf{16.84}\\
		\hline
		A.I.&25.00&24.15&21.75&\textbf{36.45}&\textbf{36.55}&\textbf{30.15}&\textbf{31.80}\\
		\bottomrule
	\end{tabular}
	
	\caption{The results of the faithfulness evaluation for LRP, CLRP, SGLRP and FG-CAM.}
	\label{tab:4}
\end{table*}

\begin{table*}
	\centering
	\begin{tabular}{lccccccc}
		\toprule
		&LRP&CLRP&SGLRP&FG-Grad-CAM&\makecell{FG-Grad-CAM\textsuperscript{D}}&FG-Score-CAM&\makecell{FG-Score-CAM\textsuperscript{D}}\\
		\midrule
		Insertion&0.55344&0.56740&0.51655&0.60757&0.61386&0.58608&0.60566\\
		\hline
		Deletion&0.09975&0.19168&0.22315&0.08668&0.11271&0.08980&0.11143\\
		\hline
		Over-all&0.45369&0.37572&0.29340&\textbf{0.52089}&\textbf{0.50115}&\textbf{0.49628}&\textbf{0.49423}\\
		\bottomrule
	\end{tabular}
	\caption{The results of insertion and deletion tests for LRP, CLRP, SGLRP and FG-CAM.}
	\label{tab:5}
\end{table*}

As shown in Table~\ref{tab:1}, FG-CAM performs better than other methods in both shallow and intermediate layers. For example, in Layer13, FG-Grad-CAM achieves 19.65 A.D. and 32.35 A.I. respectively, which outperforms other methods by a large margin. Therefore, the fine-grained explanations generated by FG-CAM can highlight important pixels in the image, and have higher faithfulness than explanations generated by other methods.
From the experimental results, it can be found that as the target layer becomes progressively shallower, the results of Score-CAM and Grad-CAM become significantly worse, while the performance of Layer-CAM declined obviously, and there is a growing gap between other methods and FG-CAM.
FG-CAM with denoising also performs better than FG-CAM in the evaluation.

To further verify the superiority of FG-CAM, we also conduct deletion and insertion tests proposed by \citeauthor{A16}, and also adopted by~\citeauthor{A5,A4,A39}.
The deletion metric measures the decrease of the output softmax value for class $c$ when important pixels are gradually removed, where the importance of pixels is given by explanation.
If the explanation indicates the importance of pixels correctly, the output softmax value should drop sharply, and the area under the deletion curve (AUC) should be low.
Similarly, the insertion metric measures the increase of the output softmax value when important pixels are gradually added, with higher AUC indicating a better explanation.

There are several methods of removing pixels \shortcite{A17}, such as setting the pixel value to zero or using blurred pixels instead, each with its own advantages and disadvantages.
In this experiment, for the deletion metric, we replaced the original pixel with the blurred pixel, and for the insertion metric, we replaced the blurred pixel with the original pixel. We removed or added 448 pixels on each step. 
We use the $Over-all$ score to comprehensively evaluate deletion and insertion results, as \citeauthor{A39}. The $Over-all$ can be calculated by $AUC(insertion)-AUC(deletion)$. 

The average result over 2000 images is reported in Table~\ref{tab:2}.
Similar to the results of the previous experiments, we find that as the target layer becomes progressively shallower, the $Over-all$ scores of Score-CAM and Grad-CAM decrease significantly, while FG-CAM is barely affected and achieves the highest $Over-all$ score.
Whether our method is used or not makes a huge difference, \eg in Layer13, FG-Grad-CAM and FG-Score-CAM have $Over-all$ scores 0.42958 and 0.25348 higher than Grad-CAM and Score-CAM, respectively. 
FG-CAM with denoising does not have much impact on the final $Over-all$ score.

\normalsize {\textbf{Localization Evaluation}}. 
We also measure the quality of explanation through localization ability.
We did not use the pointing game proposed by \citeauthor{A18} for this measure because it is too simple, requiring only the point with max value in the explanation to fall into the target object bounding box, which cannot distinguish each method significantly. 
Following \citeauthor{A4}, we measure how many values in the explanation fall into the target object bounding box, that is, measure the ability of explanation to locate target object using the $Proportion$ metric, where $Proportion=\frac{\sum L_{(i,j)\in bbox}^c}{\sum L_{(i,j)\in bbox}^c+\sum L_{(i,j)\notin bbox}^c}$. 
As \citeauthor{A4}, we randomly select 500 images with only one bounding box and no more than 50$\%$ of the pixels occupied by the box for the experiment, and observe average $Proportion$ value.

The result is reported in Table~\ref{tab:3}. 
Even though the evaluation measures only how much attention falls within the target object bounding box (it cannot accurately reflect whether the explanation fits the edge of objects), FG-CAM performs better than other methods.
On average, over 50$\%$ of the values in the explanation generated by FG-CAM fall into the target object bounding box, indicating that the explanations generated by FG-CAM are less noisy and more focused on the target object. 
It can be found that FG-CAM with denoising performs significantly better than the original FG-CAM.

All experimental results show that there is a significant difference in whether our method is used or not. 
At the same resolution, FG-CAM outperforms other methods by a large margin.
This fully verifies that FG-CAM solves the problem that CAM-based methods cannot generate fine-grained explanations while achieving generality, and has better results than Layer-CAM, which is designed for shallow layers. 
Moreover, it can be concluded from the experimental results that FG-CAM's performance is almost unaffected by the explanation resolution.
We also find that FG-Grad-CAM perform better than FG-Score-CAM on faithfulness evaluation.
We believe that this is because the global weights obtained by Grad-CAM are more accurate and using interpolation masks this advantage.

\begin{figure}[!h]
	\centering
	\includegraphics[width=1.0\linewidth]{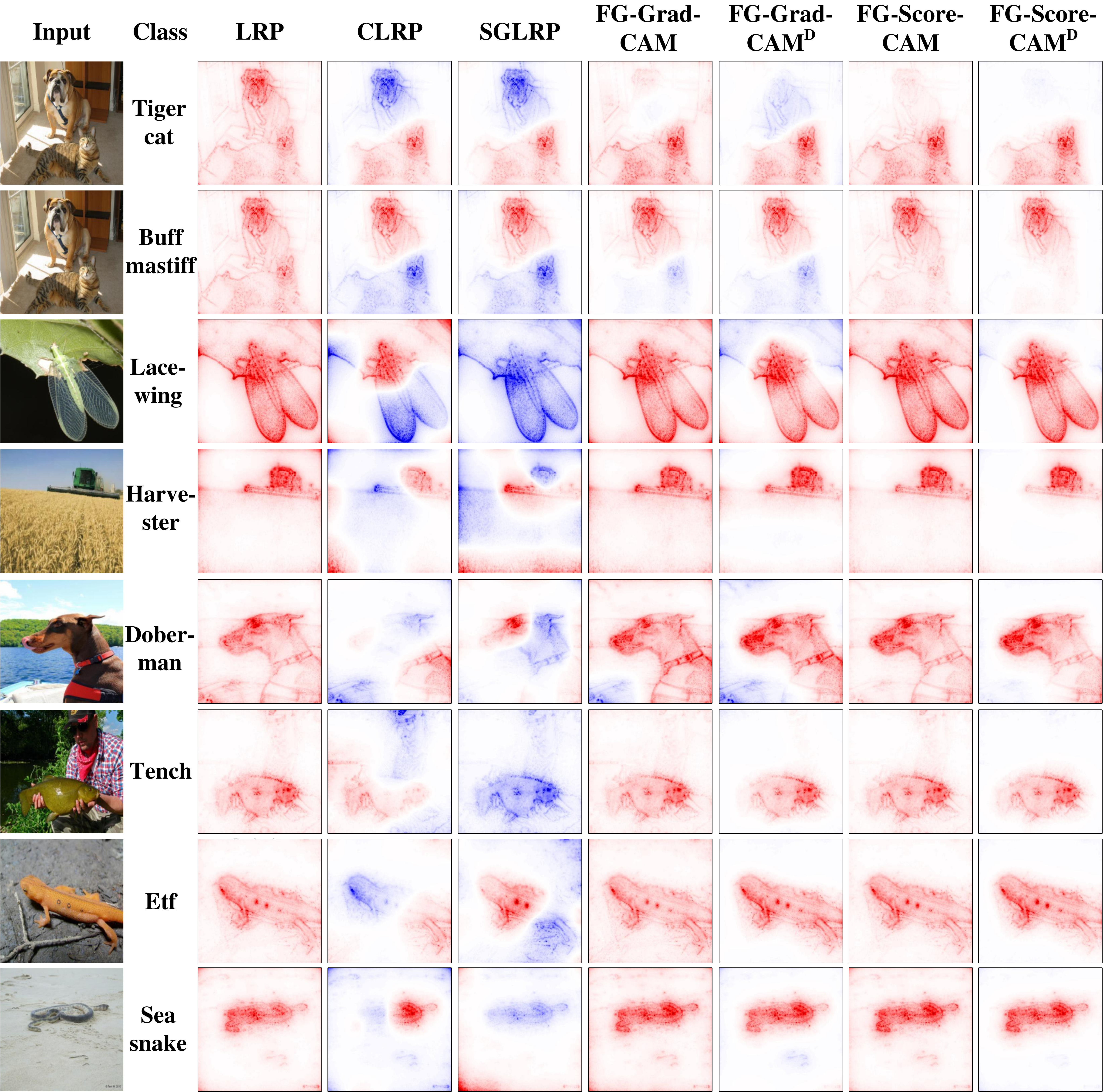}
	
	\caption{Visualization results of LRP, CLRP, SGLRP, and FG-CAM. Red indicates positive contribution pixels (the darker the color, the greater the contribution) and blue indicates negative contribution pixels.}
	\label{fig:6}
\end{figure}

\subsection{Comparison with LRP and Its Variants}

\begin{table*}[!h]
	\centering
	\begin{tabular}{lccccccc}
		\toprule
		&LRP&CLRP&SGLRP&FG-Grad-CAM&\makecell{FG-Grad-CAM\textsuperscript{D}}&FG-Score-CAM&\makecell{FG-Score-CAM\textsuperscript{D}}\\
		\midrule
		Proportion&0.48385&0.59515&0.52136&\textbf{0.49503}&\textbf{0.65496}&\textbf{0.54000}&\textbf{0.66983}\\
		\bottomrule
	\end{tabular}
	\caption{The results of evaluation on localization for LRP, CLRP, SGLRP and FG-CAM.}
	\label{tab:6}
\end{table*}

In this section, we compare LRP, SGLRP, CLRP with FG-CAM to verify that FG-CAM can generate explanations with high faithfulness and same resolution as the input image in the input layer, and outperforms these methods significantly.
We performed the same experiment. However, in the visualization and faithfulness evaluations, the explanations include negative contribution pixels to get more accurate results.

The visualization results are shown in Figure~\ref{fig:6}.
It can be found that the explanations generated by LRP are not class-discriminative (\eg on the first image, LRP even considers the pixels in the ’dog’ region to be more important).
CLRP and SGLRP often remove the important regions of the target class, and even regard pixels of important regions as negatively contributing. 
SGLRP, in particular, may regard all pixels of the target object as negatively contributing, which may make people mistakenly think that the model does not pay attention to the target object. 
FG-CAM correctly removed pixels of non-target classes without removing important parts, and focused attention on the target class. 
The explanations generated by FG-CAM with denoising focus more attention on the target class and remove a lot of noise.

For the Average Drop/Increase metric, if the proportion of positive-contributing pixels does not exceed 50\%, then we will retain only the positive-contributing pixels.
This is to ensure that only pixels with positive contribution are used. 
The result is reported in Table~\ref{tab:4}.
The A.D. and A.I. values of CLRP and SGLRP are very poor, due to the fact that they often remove important regions of the target class. 
FG-CAM is significantly better than other methods in every metric, which indicates that FG-CAM still maintains high faithfulness of explanation in the input layer and can better highlight important pixels related to the model's decision. 

The results of the deletion and insertion tests are shown in Table~\ref{tab:5}.
In this test, FG-CAM still outperforms other methods. 
For example, FG-Grad-CAM's $Over-all$ score is 0.0672 higher than LRP and 0.22749 higher than SGLRP., which is a huge lead. 
This indicates that the explanation with the same resolution as the input image generated by FG-CAM can fit the importance distribution of pixels well.

The localization evaluation results are shown in Table~\ref{tab:6}.
It can be found that CLRP and SGLRP have higher $Proportion$ value than LRP, but lower faithfulness in explanation than LRP, which further suggests that they only focus on some regions of target class and often remove important regions.
The $Proportion$ value of FG-CAM with denoising is significantly higher than that of others.
FG-CAM is also better than LRP.
This indicates that FG-CAM maintains high attention to the target class without over-removing pixels.

The experiments in this section show that FG-CAM can not only generate explanations with the same resolution as the input image, but also outperform LRP and its variants by a large margin. 
Therefore, we believe that FG-CAM greatly mitigates the shortcoming of CAM-based methods.

\section{Conclusion}
In this paper, we propose FG-CAM, a novel explanation method.
FG-CAM uses the relationship between two adjacent layers of feature maps with resolution difference to gradually increase the explanation resolution, while finding the contributing pixels and filtering out the pixels that do not contribute. 
FG-CAM solves the problem that CAM-based methods cannot generate fine-grained explanations from another perspective. 
It not only maintains the characteristics of CAM-based methods, but also produces better results. 	
Moreover, the performance of FG-CAM is almost unaffected by the explanation resolution.
FG-CAM greatly outperforms CAM-based methods in both shallow and intermediate layers, and outperforms LRP and its variants in the input layer significantly.
We also present FG-CAM with denoising, which is able to generate less noisy explanations with almost no change in explanation faithfulness. 
Our method provides a novel idea to solve the shortcoming of CAM methods, and achieves very good results. 
We believe that it will stimulate more research in this area.

\section{Acknowledgments}
This work was funded by National Natural Science Foundation
of China (No. 62272045).

\bibliography{aaai24}

\end{document}